\pdfoutput=1

\documentclass[11pt]{article}

\usepackage[final]{acl}

\usepackage{times}
\usepackage{latexsym}

\usepackage[T1]{fontenc}

\usepackage[utf8]{inputenc}

\usepackage{microtype}
\usepackage{inconsolata}
\usepackage{algorithm}
\usepackage{algorithmic}
\usepackage{booktabs}
\usepackage{xcolor}
\usepackage{amssymb} 
\usepackage{amsmath}
\usepackage{dsfont}
\usepackage{multirow} 
\usepackage{multicol} 
\usepackage{times}  
\usepackage{helvet}  
\usepackage{courier}  
\usepackage{graphicx} 
\usepackage{natbib}  
\usepackage{caption} 
\usepackage{enumitem}

\title{Evaluating, Understanding, and Improving Constrained Text Generation for Large Language Models}

\author{Xiang Chen \and Xiaojun Wan \\
        Wangxuan Institute of Computer Technology, Peking University \\
Center for Data Science, Peking University 
\\ 
The MOE Key Laboratory of Computational Linguistics, Peking University \\
\texttt{\{caspar,wanxiaojun\}@pku.edu.cn}}

\begin{document}
\maketitle
\begin{abstract}
Advancements in natural language generation (NLG) and large language models (LLMs) have led to proficient text generation in various tasks. However, integrating intricate constraints into neural text generation, due to LLMs' opacity, remains challenging. This study investigates constrained text generation for LLMs, where predefined constraints are applied during LLM's generation process. Our research mainly focuses on mainstream open-source LLMs, categorizing constraints into lexical, structural, and relation-based types. We also present various benchmarks to facilitate fair evaluation. The study addresses some key research questions, including evaluating, understanding and improving constrained text generation for LLMs. Results illuminate LLMs' capacity and deficiency to incorporate constraints and provide insights for future developments in constrained text generation. Codes and datasets will be released upon acceptance.
\end{abstract}

\section{Introduction}

Recent advances in the field of natural language generation (NLG) and large language models (LLMs)~\cite{zhao2023survey} have resulted in models able to produce realistic, coherent, and fluent texts in a multitude of natural language processing tasks. However, it is still challenging to incorporate complex constraints into neural text generation during the generation process due to the black-box nature of the LLMs. Thus, the \textit{constrained text generation}, which aims to force the LLMs to satisfy some pre-specified constraints, may be an important research topic towards better-controlled behaviours of LLMs. In addition, contrained text generation may improve the performance of many downstream tasks~\cite{hokamp2017lexically,post2018fast,lu2021neurologic,lu-etal-2022-neurologic}, including machine translation, recipe generation, dialogue response generation and table-to-text generation.

In this paper, we conduct a thorough analysis of constrained text generation on various open-source LLMs, including LLaMA-2~\cite{touvron2023llama} and Mistral~\cite{jiang2023mistral}. We also evaluate ChatGPT~\cite{chatgpt} and GPT-4~\cite{openai2023gpt4} for comparison. In order to evaluate the multifaceted generative capabilities of these LLMs, we devised the following three categories of constraints with varying levels of complexity:
\begin{itemize}[leftmargin=1em]
    \item \textbf{Lexical Constraint}~\cite{hokamp2017lexically,post2018fast,hu2019improved,chen2021lexical}: Given a set of keywords, the output text of the model is mandated to incorporate these designated keywords. These keywords can be based on grammatical rules, semantic restrictions, or stylistic guidelines.
    \item \textbf{Structural Constraint}~\cite{wang2021neural,lu2023bounding}: Regarding constraints pertaining to the structure of the output text, they encompass aspects such as sentence counts, word counts, and other related factors. 
    \item \textbf{Relation Constraint}~\cite{NEURIPS2022_ab63a1a3,bastan-etal-2023-neurostructural}: Given the relation triplets (\textit{head, relation, tail}) and force the model output to include these relation constraints. 
\end{itemize}

\begin{table*}[t!]
    \centering
    \small
    
    \begin{tabular}{l|c|p{0.6\textwidth}}
        \toprule
        Constraints & Type & Prompt\\
        \midrule
        \texttt{Keyword}$(w_1,..., w_n)$ & Lexical & Generate a sentence with keywords: ``improvise", ``barrel", ``transport", ``work", ``tool" \\
        \texttt{Order}$(w_i, w_j)$ & Structural & Generate a sentence which contains ``walk'' and ``house'', the word ``walk'' must come before ``house'' in the sentence. \\
        \texttt{WordCount}$(l)$ & Structural & Generate a sentence with exactly 10 words. \\
        \texttt{InSen}$(w, y^k)$ & Structural & Generate a story where the 2nd sentence of the story must contain the word ``cat''. \\
        \texttt{SentCount}$(l)$ & Structural & Generate a paragraph with exactly 5 sentences. \\
        \texttt{Rel}$(h, r, t)$ & Relation & Generate a sentence with keywords: ``way'' and ``worked''. The dependency relation between ``way'' and ``worked'' is ``relative clause modifier''. \\
        \bottomrule
    \end{tabular}
    \caption{The definition of constraints and the prompts used in this study.}
    \label{tab: constraints}
\end{table*}

Table~\ref{tab: constraints} shows the definition and prompts of all the constraint types. In pursuit of a more equitable and precise evaluation, we create datasets for each of the aforementioned three categories. 
Based on the datasets for evaluation, we conduct extensive experiments aimed at investigating the following research questions (RQs):

\begin{itemize}[leftmargin=1em]
    \item \textbf{RQ1: Evaluating Constrained Text Generation}: To what extent do existing LLMs address the textual constraints? How about the performance gap between the open-source and close-source LLMs?
    \item \textbf{RQ2: Understanding Constrained Text Generation}: How to understand and explain the constrained text generation capacity of LLMs?
    \item \textbf{RQ3: Improving Constrained Text Generation}: How can the constrained text generation capacity be further improved, especially for the open-source LLMs?
\end{itemize}


To explore these research questions, we initially assessed the constrained text generation capabilities of various LLMs and observed significant performance disparities between open-source LLMs and GPTs. Based on these experimental findings, we conducted a more in-depth analysis. Specifically, we employed methods such as consistency calculations, probing, and saliency score analysis to scrutinize the mechanisms and reasons behind the failure of LLMs in constrained text generation. Furthermore, based on the aforementioned analysis, we propose a simple plug-and-play attention reweighting method that enhances the constrained text generation capabilities of open-source LLMs. We believe that our experimental outcomes and proposed approach may offer valuable insights for subsequent investigations in the realm of constrained text generation.

\begin{table*}[t!]
    \centering
    \small
    \begin{tabular}{l|cc|cc|c}
        \toprule
        \textbf{Model Name} & \textbf{Accuracy} $\uparrow$ & \textbf{Coverage} $\uparrow$ & \textbf{BLEU-4} $\uparrow$ & \textbf{ROUGE-L} $\uparrow$ & \textbf{PPL} $\downarrow$ \\
        \midrule
        LLaMA2-7B-Chat & 82.15 ($\pm$0.93) & 94.69 ($\pm$0.25) & 10.60 ($\pm$0.16) & 22.78 ($\pm$0.12) & \color{blue}{46.28 ($\pm$0.49)} \\
        LLaMA2-13B-Chat & \color{blue}{84.17 ($\pm$0.47)} & \color{blue}{95.51 ($\pm$0.15)} & 10.27 ($\pm$0.13) & 22.80 ($\pm$0.07) & 48.66 ($\pm$0.35) \\
        Vicuna-7B & 72.61 ($\pm$1.03) & 91.30 ($\pm$0.22) & 7.94 ($\pm$0.24) & 20.71 ($\pm$0.15) & 71.90 ($\pm$14.83) \\
        Vicuna-13B & 74.22 ($\pm$1.08) & 92.14 ($\pm$0.50) & 10.65 ($\pm$0.15) & 22.82 ($\pm$0.17) & 55.91 ($\pm$2.44) \\
        Mistral-7B-Instruct & 80.35 ($\pm$0.59) & 95.03 ($\pm$0.19) & 12.21 ($\pm$0.27) & 23.17 ($\pm$0.06) & 50.59 ($\pm$0.86) \\
        Falcon-7B-Instruct & 55.34 ($\pm$1.55) & 87.27 ($\pm$0.63) & \color{blue}{15.90 ($\pm$0.42)} & \color{blue}{24.97 ($\pm$0.29)} & 88.69 ($\pm$6.88) \\
        \midrule
        GPT-3.5 & 94.86 & 98.69 & 16.10 & 25.75 & \textbf{45.37}\\
        GPT-4 & \textbf{97.26} & \textbf{99.33} & \textbf{16.49} & \textbf{25.98} & 52.23 \\
        \bottomrule
    \end{tabular}
    \caption{Evaluation results for lexical constraint. Bold numbers represent the highest metrics, and blue numbers indicate the highest metrics among the open-source LLMs. For open-source LLMs, the results are sampled over 5 runs with different random seeds. For GPT-3.5 and GPT-4, the results are obtained by a single run with the temperature setting to zero due to limited computational resources.}
    \label{tab:keyword}
\end{table*}

\begin{table*}[t!]
\centering
\resizebox{2.05\columnwidth}{!}{
\begin{tabular}{l|ccccc|cc}
\toprule
\multirow{2}{*}{\textbf{Model Name}} & \multicolumn{5}{c|}{\texttt{InSen}} & \multicolumn{2}{c}{\texttt{Order}} \\
& \textbf{Accuracy} & ($\pm1$) & ($\pm2$) & ($\pm3$) & \textbf{PPL} $\downarrow$ & \textbf{Accuracy} & \textbf{PPL} $\downarrow$ \\
\midrule
LLaMA2-7B-Chat & 34.96 ($\pm$1.10) & 56.98 ($\pm$0.73) & 69.10 ($\pm$1.06) & 76.02 ($\pm$0.79) & 17.83 ($\pm$0.10) & 45.50 ($\pm$0.64) & 90.36 ($\pm$3.07) \\
LLaMA2-13B-Chat & \color{blue}{37.16 ($\pm$0.15)} & \color{blue}{62.36 ($\pm$1.17)} & \color{blue}{74.10 ($\pm$1.10)} & \color{blue}{80.58 ($\pm$0.82)} & 16.03 ($\pm$0.12) & 51.34 ($\pm$0.84) & 88.44 ($\pm$4.22) \\
Vicuna-7B & 19.44 ($\pm$0.72) & 39.48 ($\pm$0.87) & 51.32 ($\pm$1.78) & 59.20 ($\pm$1.71) & 15.15 ($\pm$0.17) & 48.90 ($\pm$1.35) & 80.93 ($\pm$2.39) \\
Vicuna-13B & 21.68 ($\pm$1.14) & 43.38 ($\pm$2.04) & 55.16 ($\pm$1.47) & 63.58 ($\pm$1.00) & \color{blue}{15.09 ($\pm$0.21)} & 47.80 ($\pm$1.21) & 84.45 ($\pm$2.91) \\
Mistral-7B-Instruct & 23.84 ($\pm$0.48) & 43.86 ($\pm$0.62) & 55.92 ($\pm$0.94) & 64.60 ($\pm$1.05) & 15.66 ($\pm$0.17) & \color{blue}{52.08 ($\pm$1.09)} & \color{blue}{64.12 ($\pm$1.49)} \\
Falcon-7B-Instruct & 16.24 ($\pm$0.60) & 31.40 ($\pm$0.78) & 41.44 ($\pm$1.76) & 50.34 ($\pm$1.45) & 42.31 ($\pm$5.68) & 44.66 ($\pm$1.10) & 206.5 ($\pm$26.23) \\
\midrule
GPT-3.5 & 37.10 & 68.40 & 79.70 & 87.20 & \textbf{14.30} & 91.40 & 72.76\\
GPT-4 & \textbf{72.40} & \textbf{89.90} & \textbf{95.40} & \textbf{95.90} & 24.48 & \textbf{92.30} & \textbf{56.82} \\
\bottomrule
\end{tabular}
}
\caption{Experiment results for \texttt{InSen} and \texttt{Order} constraints.}
\label{tab: struc}
\end{table*}

\section{Task Definition} 
\subsection{Lexical Constraint} The input of a lexical constraint \texttt{Keyword}$(w_1,w_2,...,w_n)$ is an unordered set of $n$ keywords $X=\{w_1,w_2,...,w_n\}$. The expected model output is a simple and fluent sentence $Y=(y_1,y_2,...,y_m)$, where the sentence $Y$ must include all of the required keywords with reasonable morphological inflections. Formally, let $I(w_i)=\{\overline{w}_i^{(1)},\overline{w}_i^{(2)},...,\overline{w}_{i}^{(n_i)}\}$ be all forms of inflections of keyword $w_i$, the output $Y$ must contain at least one of these inflections for every keyword:
\begin{equation}
    \forall w_i \in X, \exists \overline{w}_i^{(j)} \in I(w_i), \overline{w}_i^{(j)} \in Y.
\end{equation}

\subsection{Structural Constraint} Following~\citet{wang2021neural}, we study the following three kinds of structural constraints in this paper:
\begin{itemize}[leftmargin=1em]
    \item \texttt{Order}$(w_i, w_j)$: the keyword $w_i$ is before $w_j$ in the sentence.
    \item \texttt{InSen}$(w, y^k)$: the keyword $w$ exists in the $k^{th}$ sentence of paragraph $y$.
    \item \texttt{WordCount}$(l)$: generate a sentence with exactly $l$ words.
    \item \texttt{SentCount}$(l)$: generate a paragraph with exactly $l$ sentences. \\
\end{itemize}

\subsection{Relation Constraint} Following~\citet{NEURIPS2022_ab63a1a3}, relation constraint \texttt{Rel}$(h,r,t)$ is constituted by the \textit{head} $h$, the \textit{tail} $t$, and the relation $r$ between them. Relation constraints necessitate the presence of both $h$ and $t$ in the model output, with their relation being defined by $r$, which can encompass a variety of arbitrarily defined relationships. In this paper, we employ the most fundamental dependency relation as the benchmark for testing.

\section{RQ1: Evaluating Constrained Text Generation}
\label{sec: rq1}

In this section, we construct benchmarks to evaluate the constrained text generation ability of LLMs. Due to the page limit, we only present the evaluation results and analysis. See Appendix~\ref{appendix: dataset} for the dataset construction process, Appendix~\ref{appendix: metric} for evaluation metrics, Appendix~\ref{appendix: experimental} for experimental details, and Appendix~\ref{appendix: case} for case study.

We choose widely-used LLMs including LLaMA2~\cite{touvron2023llama}, Vicuna~\cite{vicuna2023}, Mistral~\cite{jiang2023mistral}, Falcon~\cite{falcon40b}, GPT-3.5~\cite{chatgpt} and GPT-4~\cite{openai2023gpt4} for the evaluation.

\subsection{Results for Lexical Constraint} 

Table~\ref{tab:keyword} shows the results for lexical constraints. GPT-4 has the highest accuracy of 97.26\% and word coverage of 99.33\%. GPT-3.5 has the second-highest accuracy of 94.86\% and word coverage of 98.69\% with better PPL than GPT-4. While slightly lower than GPT-4, it still demonstrates a strong ability to satisfy the lexical constraints. Among the open-source LLMs, LLaMA2-13B-Chat achieves the best accuracy and coverage. Falcon-7B-Instruct achieves the best BLEU-4 and ROUGE-L.




\subsection{Results for Structural Constraint}

\begin{table*}[t!]
\centering
\small 
\begin{tabular}{l|cccc|c}
\toprule
\textbf{Model Name} & \textbf{Pearson} $\uparrow$ & \textbf{Kendall-Tau} $\uparrow$ & \textbf{Accuracy} $\uparrow$ & \textbf{MAE} $\downarrow$ & \textbf{PPL} $\downarrow$ \\
\midrule
LLaMA2-7B-Chat & 0.683 ($\pm$0.005) & 0.569 ($\pm$0.003) & 5.02 ($\pm$0.13) & 8.10 ($\pm$0.10) & 179.75 ($\pm$1.10) \\
LLaMA2-13B-Chat & \color{blue}{0.797 ($\pm$0.009)} & \color{blue}{0.682 ($\pm$0.009)} & 6.10 ($\pm$0.64) & 7.14 ($\pm$0.06) & 174.00 ($\pm$1.11) \\
Vicuna-7B & 0.131 ($\pm$0.018) & 0.186 ($\pm$0.018) & 3.36 ($\pm$0.37) & 12.75 ($\pm$0.43) & \color{blue}{95.84 ($\pm$2.17)} \\
Vicuna-13B & 0.193 ($\pm$0.083) & 0.262 ($\pm$0.030) & 4.86 ($\pm$0.47) & 8.66 ($\pm$0.45) & 209.05 ($\pm$6.88) \\
Mistral-7B-Instruct & 0.593 ($\pm$0.269) & 0.593 ($\pm$0.011) & \color{blue}{8.26 ($\pm$1.18)} & \color{blue}{5.60 ($\pm$0.35)} & 166.37 ($\pm$7.33) \\
Falcon-7B-Instruct & 0.326 ($\pm$0.226) & 0.374 ($\pm$0.008) & 1.60 ($\pm$0.16) & 11.70 ($\pm$0.31) & 807.49 ($\pm$50.77)\\
\midrule
GPT-3.5 & 0.986 & 0.942 & 33.80 & 1.14 & \textbf{59.47} \\
GPT-4 & \textbf{0.994} & \textbf{0.973} & \textbf{50.80} & \textbf{0.60} & 102.92 \\
\bottomrule
\end{tabular}
\caption{Experiment results for \texttt{WordCount} constraint.}
\label{tab: wordcount}
\end{table*}

\begin{table*}[t!]
\centering
\small 
\begin{tabular}{l|cccc|c}
\toprule
\textbf{Model Name} & \textbf{Pearson} $\uparrow$ & \textbf{Kendall-Tau} $\uparrow$ & \textbf{Accuracy} $\uparrow$ & \textbf{MAE} $\downarrow$ & \textbf{PPL} $\downarrow$\\
\midrule
LLaMA2-7B-Chat & \color{blue}{0.748 ($\pm$0.013)} & 0.707 ($\pm$0.009) & 23.92 ($\pm$0.72) & \color{blue}{2.93 ($\pm$0.06)} & 17.86 ($\pm$0.09) \\
LLaMA2-13B-Chat & 0.720 ($\pm$0.008) & \color{blue}{0.737 ($\pm$0.006)} & \color{blue}{29.58 ($\pm$0.87)} & 4.95 ($\pm$0.11) & \color{blue}{13.15 ($\pm$0.05)} \\
Vicuna-7B & 0.380 ($\pm$0.027) & 0.303 ($\pm$0.017) & 7.78 ($\pm$0.74) & 5.11 ($\pm$0.09) & 17.85 ($\pm$7.55) \\
Vicuna-13B & 0.431 ($\pm$0.017) & 0.329 ($\pm$0.015) & 8.46 ($\pm$0.32) & 5.97 ($\pm$0.10) & 13.25 ($\pm$0.33) \\
Mistral-7B-Instruct & 0.381 ($\pm$0.040) & 0.316 ($\pm$0.024) & 11.58 ($\pm$0.39) & 5.04 ($\pm$0.11) & 15.36 ($\pm$0.12) \\
Falcon-7B-Instruct & 0.159 ($\pm$0.020) & 0.182 ($\pm$0.013) & 6.04 ($\pm$0.46) & 7.19 ($\pm$0.67) & 87.82 ($\pm$24.85) \\
\midrule
GPT-3.5 & \textbf{0.946} & \textbf{0.834} & \textbf{33.40} & \textbf{1.35} & \textbf{11.19} \\
GPT-4 & 0.929 & 0.802 & 28.50 & 1.53 & 24.59 \\
\bottomrule
\end{tabular}
\caption{Experiment results for \texttt{SentCount} constraint.}
\label{tab: sentcount}
\end{table*}

\paragraph{Poor ability of sentence positioning for \texttt{InSen}, except for GPT-4.} 

Firstly, we analyze the fulfillment of the \texttt{InSen} constraints, and from the experiments, we observe that the challenge in satisfying this constraint lies not in inserting the keyword $w$ into a sentence but rather in determining which sentence should be the $k^{th}$ one to be inserted. In the subsequent paragraphs, we shall refer to this ability of the model as ``\textit{sentence positioning}".

Table~\ref{tab: struc} shows that all the LLMs tested in our study exhibited comparatively lower accuracy in adhering to the \texttt{InSen} constraints except for GPT-4. LLaMA2-13B-Chat exhibits the highest accuracy of 37.16\% among all the open-source LLMs, which is competitive with the 37.10\% of GPT-3.5.

Surprisingly, compared to GPT-3.5, GPT-4 exhibits a significant improvement in sentence positioning ability, achieving an accuracy of 72.40\%, and an impressive accuracy of 89.90\%, 95.40\%, and 95.90\%, respectively. 
We hypothesize that this enhancement may stem from GPT-4's improved reasoning capabilities, enabling it to handle sentence counting more effectively.




\paragraph{Still large gap between open-source LLMs and close-source LLMs for other structural constraints.} 
In the context of \texttt{Order} constraint, Table~\ref{tab: struc} shows that GPT-4 achieves an accuracy of 92.30\%, and GPT-3.5 achieves a competitive performance of 91.40\%. These results surpass all other tested open-source LLMs by a large margin. For other open-source LLMs, we observed that the accuracy fluctuates around a 50\% random baseline, indicating that these LLMs exhibit minimal capability in handling word order constraints. Similar observations also apply to \texttt{WordCount} (Table~\ref{tab: wordcount}) and \texttt{SentCount} (Table~\ref{tab: sentcount}) constraints,



\begin{table}[t!]
\centering
\small 
\begin{tabular}{l|cc}
\toprule
\textbf{Model Name} & \textbf{Accuracy} $\uparrow$ & \textbf{PPL} $\downarrow$\\
\midrule
LLaMA2-7B-Chat    & 29.14 ($\pm$0.95) & 47.66 ($\pm$3.87) \\
LLaMA2-13B-Chat   & \color{blue}{35.46 ($\pm$0.24)} & \color{blue}{45.66 ($\pm$0.32)} \\
Vicuna-7B         & 19.80 ($\pm$0.45) & 79.06 ($\pm$0.00) \\
Vicuna-13B        & 23.41 ($\pm$0.51) & 58.45 ($\pm$0.35) \\
Mistral-7B-Instruct & 25.41 ($\pm$0.40) & 57.46 ($\pm$8.82) \\
Falcon-7B-Instruct & 17.18 ($\pm$0.92) & 235.04 ($\pm$0.00) \\
\midrule
GPT-3.5 & 46.96 & 58.78  \\
GPT-4 & \textbf{49.48} & 62.64 \\
\bottomrule
\end{tabular}
\caption{Experiment results for relation constraint.}
\label{tab: relation}
\end{table}

\subsection{Results for Relation Constraint}
\paragraph{Close-source LLMs outperform open-source LLMs, albeit by a narrow margin.} 

From the results in Table~\ref{tab: relation}, GPT-4 achieves the best accuracy of 49.48\% by sacrificing a certain extent of PPL, which underscores that LLMs like GPT-4 can leverage their extensive corpora to learn intricate syntactic relationships between words. However, the performance discrepancy between close-source and open-source LLMs is smaller than the previous constraints. LLaMA2-13B-Chat reaches an accuracy of 35.46\%, which is competitive to GPT-4.

\paragraph{Relation constraint is still challenging.} In summary, we contend that the incorporation of relation constraints remains notably challenging for current state-of-the-art LLMs. The relative improvements achieved by the models are generally modest, and certain categories of relation constraints exhibit remarkably low accuracy. Despite the ability of existing LLMs to generate coherent text, the question of whether they genuinely comprehend human grammar rules looms large. This aspect underscores one of the potential directions for future research.

\begin{figure*}[t!]
    \centering
    \includegraphics[width=0.98\textwidth]{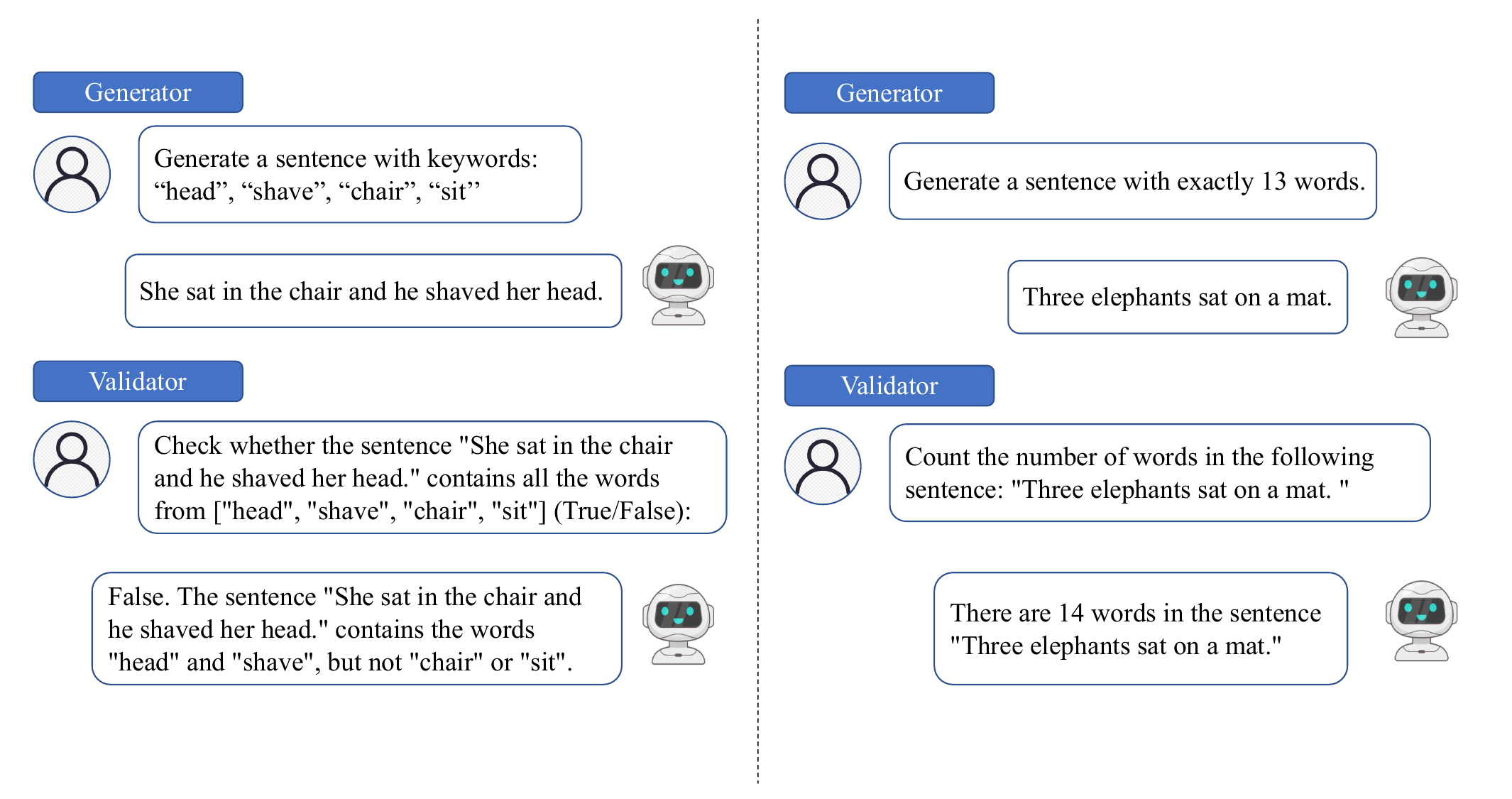}
    \caption{Examples of constructing prompts to evaluate the constraint consistency.}
    \label{fig: GV}
\end{figure*}

\section{RQ2: Understanding Constrained Text Generation}
\label{sec: rq2}

In Section~\ref{sec: rq1}, we evaluate and analyse the performance of various constrained text generation tasks across different LLMs. However, the reasons behind the success or failure of LLMs are still unknown. In this section, we employ the following three approaches to delve deeper into the mechanisms behind LLMs in fulfilling the constraints:

\begin{itemize}[leftmargin=1em]
    \item Investigating consistency (Section~\ref{ssec: cons}): to validate whether LLMs \textbf{understand} the constraints.
    \item Probing the hidden states (Section~\ref{ssec: probing}): to investigate why LLMs \textbf{can} satisfy the constraints.
    \item Saliency score calculations (Section~\ref{ssec: attention}): to investigate why LLMs \textbf{can not} satisfy the constraints.
\end{itemize}

\subsection{Constraint Consistency}
\label{ssec: cons}

For constrained text generation, LLMs may merely output sentences that comply with constraints based on their pretraining corpus, without truly understanding the meaning of the constraints. In this section, we adopt the idea of Generator-Validator Consistency~\cite{li2023benchmarking} (GV-consistency) to detect whether the LLMs truly understand the constraints. GV-consistency is an issue that LLMs can correctly generate the answers, but may not validate whether the given answers are correct or not. LLMs may also suffer from this type of inconsistency in constrained text generation. Therefore, we can further query the LLMs to validate whether the constraints are satisfied to investigate the inconsistency. We denote the consistency as ``Constraint Consistency'' in the following paragraphs.

\paragraph{Validator Construction} We first construct the validators to query the LLMs. For each constraint $x$, LLMs generate the sentences or the document $g(x)$. We use $g(x)$ as inputs, asking LLMs to check whether $g(x)$ satisfies the constraint $x$. LLMs give the validator response $v(x, g(x))$. Figure~\ref{fig: GV} illustrates two types of validators for different constraints. For lexical constraints, \texttt{InSen}, \texttt{Order} and relation constraints, we query the LLMs to obtain a true-or-false response about whether the constraints are satisfied. In these cases, $v(x, g(x))\in\{0, 1\}$. For \texttt{WordCount} and \texttt{SentCount}, we query the LLMs to count the number of words or sentences. In these cases, $v(x, g(x))\in\mathbb{N}^+$.

\paragraph{Definition of Constraint Consistency} Given the constraint $x$ and generator output $g(x)$, we can automatically validate whether $g(x)$ satisfies $x$ by a program, which serves as the ground truth label for validators. We denote the ground truth as $f(x, g(x))$. Therefore, we can define the constraint consistency as follows:
\begin{itemize}[leftmargin=1em]
    \item For $v(x, g(x))\in\{0, 1\}$, constraint consistency is the F1-score between the $v(x, g(x))$ and $f(x, g(x))$.
    \item For $v(x, g(x))\in\mathbb{N}^+$, constraint consistency is the pearson correlation between the $v(x, g(x))$ and $f(x, g(x))$.
\end{itemize}

\begin{table*}[t!]
\centering
\resizebox{2.05\columnwidth}{!}{
\begin{tabular}{l|c|cccc|c}
\toprule
\textbf{Model Name} & \texttt{Keyword} & \texttt{InSen} & \texttt{Order} & \texttt{WordCount} & \texttt{SentCount} & \texttt{Rel}\\
\midrule
LLaMA2-7B-Chat & 90.14 ($\pm$0.65) & 51.90 ($\pm$1.48) & 61.42 ($\pm$0.73) & 99.58 ($\pm$0.09) & 77.04 ($\pm$0.79) & 44.42 ($\pm$1.13) \\
LLaMA2-13B-Chat & \textbf{90.40 ($\pm$0.22)} & \textbf{52.10 ($\pm$1.34)} & \textbf{68.21 ($\pm$0.48)} & \textbf{99.75 ($\pm$0.07)} & \textbf{90.69 ($\pm$0.56)} & \textbf{52.37 ($\pm$0.33)} \\
Vicuna-7B & 4.68 ($\pm$0.70) & 18.73 ($\pm$3.33) & 3.33 ($\pm$1.83) & 24.87 ($\pm$10.68) & 11.30 ($\pm$1.61) & 1.63 ($\pm$0.41) \\
Vicuna-13B & 82.27 ($\pm$0.95) & 36.79 ($\pm$1.95) & 58.99 ($\pm$1.81) & 34.27 ($\pm$20.89) & 17.85 ($\pm$2.70) & 39.73 ($\pm$0.71) \\
Mistral-7B-Instruct & 89.38 ($\pm$0.32) & 38.84 ($\pm$0.66) & 67.26 ($\pm$1.03) & 63.39 ($\pm$28.66) & 34.78 ($\pm$7.44) & 40.79 ($\pm$0.65)\\
Falcon-7B-Instruct & 71.22 ($\pm$1.33) & 27.59 ($\pm$0.87) & 60.09 ($\pm$1.18) & 30.73 ($\pm$27.94) & -6.38 ($\pm$4.43) & 29.38 ($\pm$1.48)\\
\bottomrule
\end{tabular}
}
\caption{Evaluation results for constraint consistency. GPT-3.5 and GPT-4 are not presented due to limited computational resources.}
\label{tab: consistency}
\end{table*}

\paragraph{Experimental Results} Table~\ref{tab: consistency} shows the results for constraint consistency. Among the open-source LLMs, we found that LLaMA2-13B-Chat achieves the best consistency over all kinds of constraints, which shows a well understanding of constraints. Vicuna-7B has very poor consistency, indicating that it may have a poor understanding of constraints. Combining Table~\ref{tab: consistency} with the previous experimental results, it is evident that there is a positive correlation between constraint consistency and the performance of the respective task. LLMs exhibiting high consistency also demonstrate higher performance in the corresponding constraint category.

\begin{figure}[t!]
    \centering
    \includegraphics[width=0.48\textwidth]{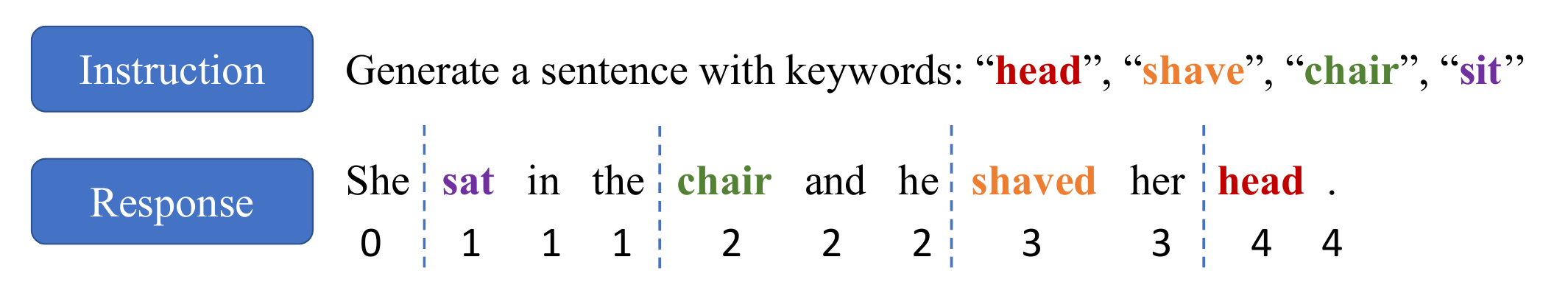}
    \caption{Illustration of label construction for the probing experiment. Each label denotes the number of completed constraints. We only label the response part.}
    \label{fig: probing}
    \vspace{-3mm}
\end{figure}

\subsection{Probing}
\label{ssec: probing}

In this section, we aim to explore the mechanisms behind why LLMs can fulfill various complex constraints successfully. It is natural to hypothesize: the existence of a \textbf{constraint completion state} mechanism within LLMs that records the status of whether the constraints have been met. This state is expected to be reflected in the hidden states of LLMs at each generation step. 

To validate this hypothesis, we employ the linear probing method to analyze the hidden states of LLMs. Linear probing is a technique used in the exploration and analysis of the internal representations of LLMs to gain insights into the model's functioning. Existing studies~\cite{tenney2018you,hewitt2019structural,vulic2020probing} mostly involve probing bidirectional language models, such as BERT~\cite{devlin2019bert}, to investigate some linguistic properties (e.g., syntax, part-of-speech tagging). In contrast to these works, our approach involves probing generative LLMs to examine the existence of constraint completion states. 

Due to the page limit, in this paper, we only validate our hypothesis using the most common lexical constraint as an example. The approach we used in this section can be easily extended to other types of constraints.

\paragraph{Method} Following the typical approach of linear probing, we train an additional linear layer on top of the last layer of LLMs, with the goal of predicting how many keywords are satisfied. We assign a label to each token of the output sequences and use the training objective of mean squared error to train the linear layer. As illustrated by Figure~\ref{fig: probing}, each label denotes the number of completed constraints. The maximum number of keywords in a sentence is set to 5 in our experiment. We compute the Pearson correlation and mean absolute error (MAE) as evaluation metrics.

\begin{table}[t!]
\centering
\small 
\begin{tabular}{l|cc}
\toprule
\textbf{Model Name} & \textbf{Pearson} $\uparrow$ & \textbf{MAE} $\downarrow$ \\
\midrule
LLaMA2-7B-Chat & 0.859 ($\pm$0.006) & 0.67 ($\pm$0.02) \\
Vicuna-7B & 0.845 ($\pm$0.008) & 0.68 ($\pm$0.03) \\
Mistral-7B-Instruct & 0.895 ($\pm$0.006) & 0.59 ($\pm$0.02) \\
Falcon-7B-Instruct & \textbf{0.898 ($\pm$0.002)} & \textbf{0.53 ($\pm$0.03)} \\
\bottomrule
\end{tabular}
\caption{Probing Results. 13B LLMs are not presented due to limited resources.}
\label{tab: probing}
\vspace{-3mm}
\end{table}

\paragraph{Experimental Results} Table~\ref{tab: probing} demonstrates the results for probing experiment. We found that for all tested open-source LLMs, the hidden states can exhibit a strong correlation (around 0.9) to the number of completed constraints, with a minimal mean absolute error smaller than 1. This experimental result indicates that hidden states have a strong predictive ability to the constraints. Therefore, LLMs may be aware of the state of constraint completion, which is stored in the hidden states.

\subsection{Saliency Score}
\label{ssec: attention}

In this section, we aim to interpret why LLMs fail to fulfill the constraints. Similar to Section~\ref{ssec: probing}, we conduct our experiment only in lexical constraint as an example. We consider using the interpretability method based on attention scores, which is a widely employed tool for explicating the performance of LLMs. We hypothesize that the failure of LLMs to fulfill the given constraints is attributed to \textbf{insufficient attention} to the keywords specified in the instructions. Therefore, we attempt to quantitatively evaluate the contribution of each word at every generation step to validate our hypothesis.

\paragraph{Method} Following the common practice, we use the saliency score method~\cite{simonyan2013deep,NEURIPS2019_2c601ad9,wang2023label} to interpret the importance of every token in the generation process:

\begin{equation}
    I_l = \sum_h \left|A_{h,l}\odot \frac{\partial \mathcal{L}(x)}{\partial A_{h,l}} \right|.
\end{equation}

In this equation, $A_{h,l}\in\mathbb{R}^{n\times n}$ denotes the attention weight matrix of the $h^{th}$ head in the $l^{th}$ layer, where $n$ is the sequence length. $\mathcal{L}(x)$ denotes the loss function (i.e., cross-entropy loss) of the next token prediction. We sum over all the attention heads to produce an overall importance matrix $I_l$ for the $l^{th}$ layer. The $(i, j)$ element of matrix $I_l$ indicates the contribution of $i^{th}$ token in generating the next token of $j^{th}$ token.

We divided each sample into three parts: the input instruction $\mathcal{S}_I$ (e.g., ``Generate a sentence with keywords:''), the keywords provided in the input $\mathcal{S}_K$, and the LLMs' response $\mathcal{S}_R(j)$ when generating $j^{th}$ token. We can define the average contribution of each part when generating $j^{th}$ token by the average value of $I_l(i, j)$:

\begin{equation}
\begin{aligned}
    C_I &= \frac{1}{|\mathcal{S}_I|}\sum_{i\in \mathcal{S}_I} I_l(i, j), \forall j, \\
    C_K &= \frac{1}{|\mathcal{S}_K|}\sum_{i\in \mathcal{S}_K} I_l(i, j), \forall j, \\
    C_R &= \frac{1}{|\mathcal{S}_R(j)|}\sum_{i\in \mathcal{S}_R(j)} I_l(i, j), \forall j.
\end{aligned}
\end{equation}

\begin{figure}[t!]
    \centering
    \includegraphics[width=0.48\textwidth]{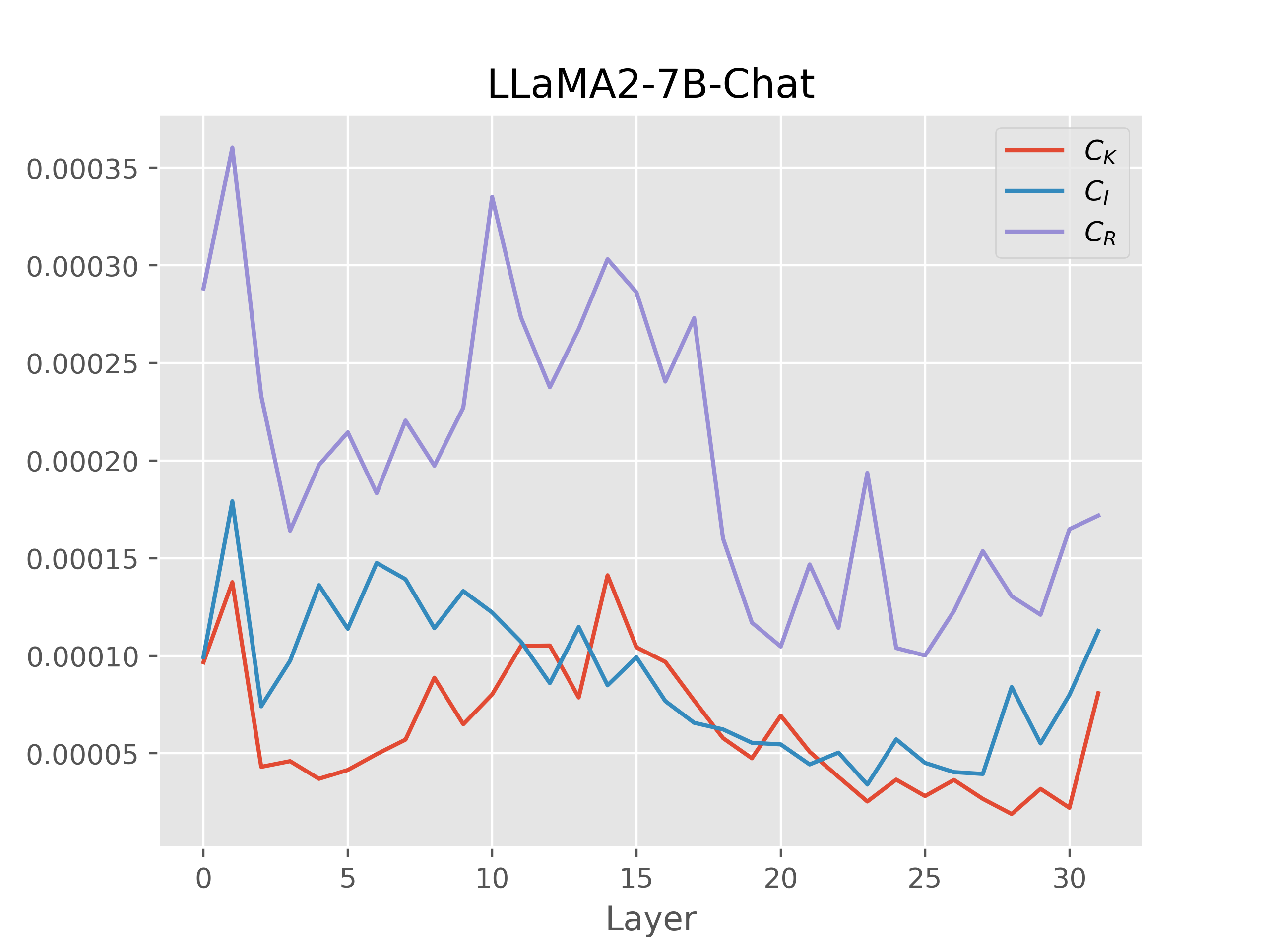}
    \caption{Results on LLaMA2-7B-Chat by layers.}
    \label{fig: llama_by_layer}
    \vspace{-3mm}
\end{figure}

\begin{figure}[t!]
    \centering
    \includegraphics[width=0.48\textwidth]{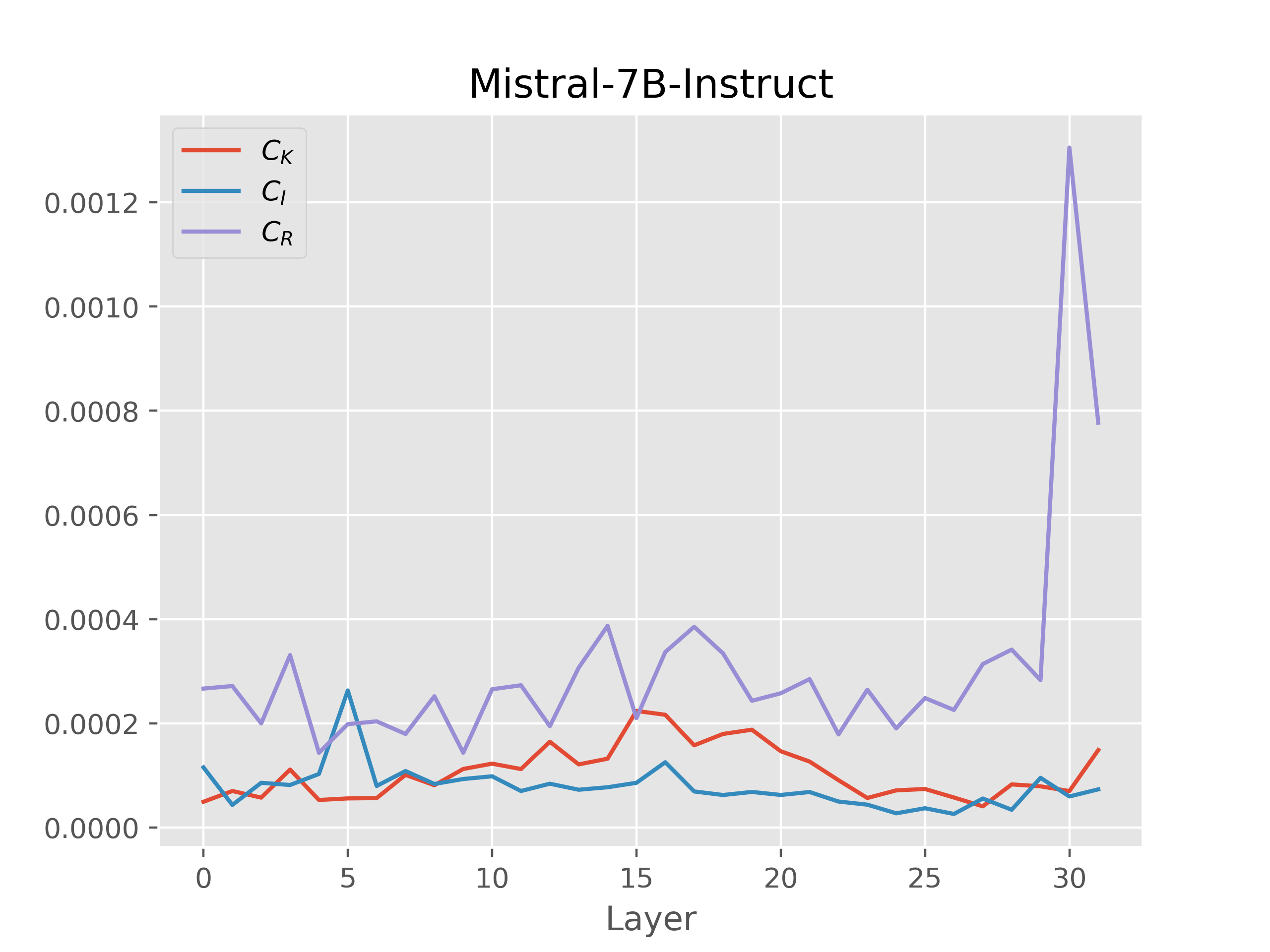}
    \caption{Results on Mistral-7B-Instruct by layers.}
    \label{fig: mistral_by_layer}
    \vspace{-3mm}
\end{figure}

\paragraph{Experimental Results} We calculate $C_I$, $C_K$ and $C_R$ by different layers on LLaMA2-7B-Chat and Mistral-7B-Instruct. Figures~\ref{fig: llama_by_layer} and~\ref{fig: mistral_by_layer} show the results. We can observe that $C_R$ is significantly larger than $C_I$ and $C_K$ across nearly all layers of both models. This phenomenon suggests that LLMs may overly focus on the content of the already generated response ($\mathcal{S}_R(j)$) during the sentence generation process, paying insufficient attention to the instruction ($\mathcal{S}_I$) and keywords ($\mathcal{S}_K$), which likely contributes to the sub-optimal effectiveness in fulfilling the constraints. Based on this observation, we can further optimize the performance of LLMs by targeting attention score designs.



\begin{table*}[t!]
    \centering
    \small
    \begin{tabular}{l|cc|cc|c}
        \toprule
        \textbf{Model Name} & \textbf{Accuracy} $\uparrow$ & \textbf{Coverage} $\uparrow$ & \textbf{BLEU-4} $\uparrow$ & \textbf{ROUGE-L} $\uparrow$ & \textbf{PPL} $\downarrow$ \\
        \midrule
        LLaMA2-7B-Chat & 82.15 ($\pm$0.93) & 94.69 ($\pm$0.25) & 10.60 ($\pm$0.16) & 22.78 ($\pm$0.12) & 46.28 ($\pm$0.49) \\
        + Re-anchoring & \textbf{88.88 ($\pm$1.36)} & \textbf{96.96 ($\pm$0.38)} & \textbf{10.73 ($\pm$0.22)} & \textbf{23.27 ($\pm$0.14)} & \textbf{44.50 ($\pm$0.64)} \\
        \midrule
        Mistral-7B-Instruct & 80.35 ($\pm$0.59) & 95.03 ($\pm$0.19) & \textbf{12.21 ($\pm$0.27)} & \textbf{23.17 ($\pm$0.06)} & \textbf{50.59 ($\pm$0.86)} \\
        + Re-anchoring & \textbf{88.98 ($\pm$0.40)} & \textbf{97.32 ($\pm$0.14)} & 11.85 ($\pm$0.31) & 23.00 ($\pm$0.16) & 51.53 ($\pm$1.42) \\
        \bottomrule
    \end{tabular}
    \caption{Evaluation results for Attention Re-anchoring.}
    \label{tab: reanchor}
    \vspace{-3mm}
\end{table*}

\section{RQ3: Improving Constrained Text Generation}
\label{sec: rq3}

In this section, we aim to explore methods for improving the constrained text generation capabilities of LLMs based on the experimental and analytical results presented earlier in Sections~\ref{sec: rq1} and~\ref{sec: rq2}. A straightforward approach involves constructing instances that satisfy constraints and fine-tuning LLMs on these instances. While this approach can evidently improve the performance on the given evaluation set, it may lack generalizability. Moreover, fine-tuning LLMs incurs substantial costs, as well as the challenge of catastrophic forgetting. Therefore, in this paper, we endeavor to investigate a simpler, easily implementable, and scalable method to enhance the constrained text generation abilities of LLMs without resorting to training. Similarly to the previous sections, we focus on the setting of lexical constraints and conduct experiments on open-source LLMs.

\subsection{Attention Re-anchoring}

In Section~\ref{ssec: attention}, we validate that LLMs have the problem of insufficient attention to the keywords in the instructions. Inspired by this, we propose a simple approach named Attention Re-anchoring to enhance the attention weights of keywords in the instruction. Specifically, we multiply a mask matrix $M$ over the query-key dot matrix of self-attention to reweight the keywords in $\mathcal{S}_K$:
\begin{equation}
\begin{aligned}
    \text{Att}(Q, K, V) = \text{softmax}\left(\frac{QK^\top\odot M}{\sqrt{d_k}} \right)V, \\
    M_{ij} = \mathds{1}[i\le j] \cdot \left\{1+\lambda\cdot\mathds{1} \left[i\in \mathcal{S}_K \right]\right\}, \forall i, j.
\end{aligned}
\end{equation}
where $\lambda$ is a hyperparameter to control the reweighting. The matrix $M$ essentially increases the weight of tokens in $\mathcal{S}_K$ on the generation process while preserving the causal mask of LLMs.

\subsection{Experimental Results}
As a preliminary attempt, we conducted experiments on LLaMA2-7B-Chat and Mistral-7B-Instruct under the setting of lexical constraints. We observed a significant accuracy improvement of up to 8\% compared to the baseline of directly employing these LLMs for inference. The PPL metric shows minimal changes, indicating that the attention re-anchoring method does not significantly affect the fluency of generated sentences.

\section{Related Work}

\paragraph{Constrained Text Generation} 
Constrained text generation allows users to generate text that meets specific criteria or adheres to certain constraints. This can be especially valuable in situations where the generated text must be finely controlled. Constrained text generation can be achieved through a variety of techniques. For lexical constraints, previous studies involved the modification of decoding algorithm~\cite{anderson2017guided,hokamp2017lexically,post2018fast,hu2019improved,mao2021extract}, sample-based generation~\cite{he2021show,miao2019cgmh,sha2020gradient}, training data augmentation~\cite{song2019code,dinu2019training,chen2021lexical} and adding additional model structure~\cite{song2020alignment,wang2021mention}. For structural constraints, \citet{wang2021neural} proposed NRETM, which equips an additional structure into various transformer-based generators to keep track of the progress of structural constraints simultaneously. For relation constraints, \citet{NEURIPS2022_ab63a1a3} created a benchmark to evaluate the language models' ability to generate sentences given keywords and their dependency relations. 

\paragraph{Evaluating Constrained Text Generation for Large Language Models} Many existing works~\cite{hendrycks2020measuring,wang2022super,li2022quantifying,zheng2023judging,wang2023pandalm} have introduced diverse methodologies and datasets aimed at probing the text generation capabilities of LLMs. However, this line of work focuses on the systematic evaluation of LLMs, not a specific kind of capability. \citet{lin2021truthfulqa} created a dataset of prompts to detect the hallucinations of GPT-3~\cite{brown2020language}. \citet{zhou2023controlled} investigated the controlled text generation of LLMs by supervised fine-tuning with natural language instructions. \citet{lu2023bounding} examined the capabilities of LLMs to generate texts with stylistic and structural prompt constraints in open-ended text generation settings. \citet{yao2023collie} constructed a benchmark to evaluate the ability of LLMs to follow the structural constraints regarding word position, length, and character counts. \citet{sun2023evaluating} tested LLMs on structural constraints and other four controlled generation tasks. \citet{zhou2023instruction} evaluated the instruction-following ability for LLMs. \citet{chen2024benchmarking} evaluated controllable text generation under diversified instruction for LLMs. 


\section{Conclusion} 

In this article, we have evaluated three categories of constraints in the domain of constrained text generation on several popular open-source LLMs and GPTs. To further understand the constrained generation process, we analyze open-source LLMs from the aspect of consistency, hidden representation and saliency score. Based on these results, we propose Attention Re-anchoring to shrink the gap between open-source LLMs and GPTs for lexical constraints.
We anticipate that the findings and analyses in this work will serve to assist and inspire future research endeavors in this field.

\section*{Limitations}
This paper primarily focuses on evaluating, understanding, and improving constrained text generation for LLMs. However, there are several limitations as follows: (1) We only evaluated three types of constraints: lexical, structural, and relation, while there exist more constraint types in practical applications. Further work could explore other types of constraints. (2) Due to page limits, Section~\ref{ssec: probing} and~\ref{ssec: attention} only analyze lexical constraints. The analytical methods and procedures may require modifications when extending to other types of constraints. Additionally, due to the black-box nature of GPT models, we could only analyze open-source LLMs. (3) The methods proposed in Section~\ref{sec: rq3} are applicable only to open-source LLMs under lexical constraint setting. It's still worth investigating whether the method can extend to other settings. Moreover, even after enhancement, the performance still falls short of that of GPT models. This may restrict potential application scenarios, necessitating further research by future scholars.



\bibliography{acl}

\begin{thebibliography}{55}
\expandafter\ifx\csname natexlab\endcsname\relax\def\natexlab#1{#1}\fi

\bibitem[{Almazrouei et~al.(2023)Almazrouei, Alobeidli, Alshamsi, Cappelli, Cojocaru, Debbah, Goffinet, Heslow, Launay, Malartic, Noune, Pannier, and Penedo}]{falcon40b}
Ebtesam Almazrouei, Hamza Alobeidli, Abdulaziz Alshamsi, Alessandro Cappelli, Ruxandra Cojocaru, Merouane Debbah, Etienne Goffinet, Daniel Heslow, Julien Launay, Quentin Malartic, Badreddine Noune, Baptiste Pannier, and Guilherme Penedo. 2023.
\newblock {Falcon-40B}: an open large language model with state-of-the-art performance.

\bibitem[{Anderson et~al.(2016)Anderson, Fernando, Johnson, and Gould}]{anderson2016spice}
Peter Anderson, Basura Fernando, Mark Johnson, and Stephen Gould. 2016.
\newblock Spice: Semantic propositional image caption evaluation.
\newblock In \emph{Computer Vision--ECCV 2016: 14th European Conference, Amsterdam, The Netherlands, October 11-14, 2016, Proceedings, Part V 14}, pages 382--398. Springer.

\bibitem[{Anderson et~al.(2017)Anderson, Fernando, Johnson, and Gould}]{anderson2017guided}
Peter Anderson, Basura Fernando, Mark Johnson, and Stephen Gould. 2017.
\newblock \href {https://doi.org/10.18653/v1/D17-1098} {Guided open vocabulary image captioning with constrained beam search}.
\newblock In \emph{Proceedings of the 2017 Conference on Empirical Methods in Natural Language Processing}, pages 936--945, Copenhagen, Denmark. Association for Computational Linguistics.

\bibitem[{Banerjee and Lavie(2005)}]{banerjee2005meteor}
Satanjeev Banerjee and Alon Lavie. 2005.
\newblock \href {https://aclanthology.org/W05-0909} {{METEOR}: An automatic metric for {MT} evaluation with improved correlation with human judgments}.
\newblock In \emph{Proceedings of the {ACL} Workshop on Intrinsic and Extrinsic Evaluation Measures for Machine Translation and/or Summarization}, pages 65--72, Ann Arbor, Michigan. Association for Computational Linguistics.

\bibitem[{Bastan et~al.(2023)Bastan, Surdeanu, and Balasubramanian}]{bastan-etal-2023-neurostructural}
Mohaddeseh Bastan, Mihai Surdeanu, and Niranjan Balasubramanian. 2023.
\newblock \href {https://doi.org/10.18653/v1/2023.acl-long.528} {{NEUROSTRUCTURAL} {DECODING}: Neural text generation with structural constraints}.
\newblock In \emph{Proceedings of the 61st Annual Meeting of the Association for Computational Linguistics (Volume 1: Long Papers)}, pages 9496--9510, Toronto, Canada. Association for Computational Linguistics.

\bibitem[{Brown et~al.(2020)Brown, Mann, Ryder, Subbiah, Kaplan, Dhariwal, Neelakantan, Shyam, Sastry, Askell et~al.}]{brown2020language}
Tom Brown, Benjamin Mann, Nick Ryder, Melanie Subbiah, Jared~D Kaplan, Prafulla Dhariwal, Arvind Neelakantan, Pranav Shyam, Girish Sastry, Amanda Askell, et~al. 2020.
\newblock Language models are few-shot learners.
\newblock \emph{Advances in neural information processing systems}, 33:1877--1901.

\bibitem[{Chen et~al.(2020)Chen, Chen, Wang, and Li}]{chen2021lexical}
Guanhua Chen, Yun Chen, Yong Wang, and Victor O.~K. Li. 2020.
\newblock \href {https://doi.org/10.24963/ijcai.2020/496} {Lexical-constraint-aware neural machine translation via data augmentation}.
\newblock In \emph{Proceedings of the Twenty-Ninth International Joint Conference on Artificial Intelligence, {IJCAI} 2020}, pages 3587--3593. ijcai.org.

\bibitem[{Chen et~al.(2022)Chen, Yang, and Wan}]{NEURIPS2022_ab63a1a3}
Xiang Chen, Zhixian Yang, and Xiaojun Wan. 2022.
\newblock \href {https://proceedings.neurips.cc/paper_files/paper/2022/file/ab63a1a325670278ba9b87fbc3e95e33-Paper-Conference.pdf} {Relation-constrained decoding for text generation}.
\newblock In \emph{Advances in Neural Information Processing Systems}, volume~35, pages 26804--26819. Curran Associates, Inc.

\bibitem[{Chen et~al.(2024)Chen, Xu, Wang, Liu, and Mao}]{chen2024benchmarking}
Yihan Chen, Benfeng Xu, Quan Wang, Yi~Liu, and Zhendong Mao. 2024.
\newblock Benchmarking large language models on controllable generation under diversified instructions.
\newblock \emph{arXiv preprint arXiv:2401.00690}.

\bibitem[{Chiang et~al.(2023)Chiang, Li, Lin, Sheng, Wu, Zhang, Zheng, Zhuang, Zhuang, Gonzalez, Stoica, and Xing}]{vicuna2023}
Wei-Lin Chiang, Zhuohan Li, Zi~Lin, Ying Sheng, Zhanghao Wu, Hao Zhang, Lianmin Zheng, Siyuan Zhuang, Yonghao Zhuang, Joseph~E. Gonzalez, Ion Stoica, and Eric~P. Xing. 2023.
\newblock \href {https://lmsys.org/blog/2023-03-30-vicuna/} {Vicuna: An open-source chatbot impressing gpt-4 with 90\%* chatgpt quality}.

\bibitem[{Devlin et~al.(2019)Devlin, Chang, Lee, and Toutanova}]{devlin2019bert}
Jacob Devlin, Ming-Wei Chang, Kenton Lee, and Kristina Toutanova. 2019.
\newblock \href {https://doi.org/10.18653/v1/N19-1423} {{BERT}: Pre-training of deep bidirectional transformers for language understanding}.
\newblock In \emph{Proceedings of the 2019 Conference of the North {A}merican Chapter of the Association for Computational Linguistics: Human Language Technologies, Volume 1 (Long and Short Papers)}, pages 4171--4186, Minneapolis, Minnesota. Association for Computational Linguistics.

\bibitem[{Dinu et~al.(2019)Dinu, Mathur, Federico, and Al-Onaizan}]{dinu2019training}
Georgiana Dinu, Prashant Mathur, Marcello Federico, and Yaser Al-Onaizan. 2019.
\newblock \href {https://doi.org/10.18653/v1/P19-1294} {Training neural machine translation to apply terminology constraints}.
\newblock In \emph{Proceedings of the 57th Annual Meeting of the Association for Computational Linguistics}, pages 3063--3068, Florence, Italy. Association for Computational Linguistics.

\bibitem[{He and Li(2021)}]{he2021show}
Xingwei He and Victor~OK Li. 2021.
\newblock Show me how to revise: Improving lexically constrained sentence generation with xlnet.
\newblock In \emph{Proceedings of AAAI}, pages 12989--12997.

\bibitem[{Hendrycks et~al.(2020)Hendrycks, Burns, Basart, Zou, Mazeika, Song, and Steinhardt}]{hendrycks2020measuring}
Dan Hendrycks, Collin Burns, Steven Basart, Andy Zou, Mantas Mazeika, Dawn Song, and Jacob Steinhardt. 2020.
\newblock Measuring massive multitask language understanding.
\newblock \emph{arXiv preprint arXiv:2009.03300}.

\bibitem[{Hewitt and Manning(2019)}]{hewitt2019structural}
John Hewitt and Christopher~D Manning. 2019.
\newblock A structural probe for finding syntax in word representations.
\newblock In \emph{Proceedings of the 2019 Conference of the North American Chapter of the Association for Computational Linguistics: Human Language Technologies, Volume 1 (Long and Short Papers)}, pages 4129--4138.

\bibitem[{Hokamp and Liu(2017)}]{hokamp2017lexically}
Chris Hokamp and Qun Liu. 2017.
\newblock \href {https://doi.org/10.18653/v1/P17-1141} {Lexically constrained decoding for sequence generation using grid beam search}.
\newblock In \emph{Proceedings of the 55th Annual Meeting of the Association for Computational Linguistics (Volume 1: Long Papers)}, pages 1535--1546, Vancouver, Canada. Association for Computational Linguistics.

\bibitem[{Hu et~al.(2019)Hu, Khayrallah, Culkin, Xia, Chen, Post, and Van~Durme}]{hu2019improved}
J.~Edward Hu, Huda Khayrallah, Ryan Culkin, Patrick Xia, Tongfei Chen, Matt Post, and Benjamin Van~Durme. 2019.
\newblock \href {https://doi.org/10.18653/v1/N19-1090} {Improved lexically constrained decoding for translation and monolingual rewriting}.
\newblock In \emph{Proceedings of the 2019 Conference of the North {A}merican Chapter of the Association for Computational Linguistics: Human Language Technologies, Volume 1 (Long and Short Papers)}, pages 839--850, Minneapolis, Minnesota. Association for Computational Linguistics.

\bibitem[{Jiang et~al.(2023)Jiang, Sablayrolles, Mensch, Bamford, Chaplot, Casas, Bressand, Lengyel, Lample, Saulnier et~al.}]{jiang2023mistral}
Albert~Q Jiang, Alexandre Sablayrolles, Arthur Mensch, Chris Bamford, Devendra~Singh Chaplot, Diego de~las Casas, Florian Bressand, Gianna Lengyel, Guillaume Lample, Lucile Saulnier, et~al. 2023.
\newblock Mistral 7b.
\newblock \emph{arXiv preprint arXiv:2310.06825}.

\bibitem[{Kingma and Ba(2015)}]{kingma2014adam}
Diederik~P. Kingma and Jimmy Ba. 2015.
\newblock \href {http://arxiv.org/abs/1412.6980} {Adam: {A} method for stochastic optimization}.
\newblock In \emph{3rd International Conference on Learning Representations, {ICLR} 2015, San Diego, CA, USA, May 7-9, 2015, Conference Track Proceedings}.

\bibitem[{Li et~al.(2022)Li, Yu, Khabsa, Zettlemoyer, Halevy, and Andreas}]{li2022quantifying}
Belinda~Z Li, Jane Yu, Madian Khabsa, Luke Zettlemoyer, Alon Halevy, and Jacob Andreas. 2022.
\newblock Quantifying adaptability in pre-trained language models with 500 tasks.
\newblock In \emph{Proceedings of the 2022 Conference of the North American Chapter of the Association for Computational Linguistics: Human Language Technologies}, pages 4696--4715.

\bibitem[{Li et~al.(2023)Li, Shrivastava, Li, Hashimoto, and Liang}]{li2023benchmarking}
Xiang~Lisa Li, Vaishnavi Shrivastava, Siyan Li, Tatsunori Hashimoto, and Percy Liang. 2023.
\newblock Benchmarking and improving generator-validator consistency of language models.
\newblock \emph{arXiv preprint arXiv:2310.01846}.

\bibitem[{Lin et~al.(2020)Lin, Zhou, Shen, Zhou, Bhagavatula, Choi, and Ren}]{lin2020commongen}
Bill~Yuchen Lin, Wangchunshu Zhou, Ming Shen, Pei Zhou, Chandra Bhagavatula, Yejin Choi, and Xiang Ren. 2020.
\newblock Commongen: A constrained text generation challenge for generative commonsense reasoning.
\newblock In \emph{Findings of the Association for Computational Linguistics: EMNLP 2020}, pages 1823--1840.

\bibitem[{Lin(2004)}]{lin2004rouge}
Chin-Yew Lin. 2004.
\newblock \href {https://aclanthology.org/W04-1013} {{ROUGE}: A package for automatic evaluation of summaries}.
\newblock In \emph{Text Summarization Branches Out}, pages 74--81, Barcelona, Spain. Association for Computational Linguistics.

\bibitem[{Lin et~al.(2021)Lin, Hilton, and Evans}]{lin2021truthfulqa}
Stephanie Lin, Jacob Hilton, and Owain Evans. 2021.
\newblock Truthfulqa: Measuring how models mimic human falsehoods.
\newblock \emph{arXiv preprint arXiv:2109.07958}.

\bibitem[{Lu et~al.(2023)Lu, Zhang, Zhang, Wang, and Yang}]{lu2023bounding}
Albert Lu, Hongxin Zhang, Yanzhe Zhang, Xuezhi Wang, and Diyi Yang. 2023.
\newblock Bounding the capabilities of large language models in open text generation with prompt constraints.
\newblock \emph{arXiv preprint arXiv:2302.09185}.

\bibitem[{Lu et~al.(2022)Lu, Welleck, West, Jiang, Kasai, Khashabi, Le~Bras, Qin, Yu, Zellers, Smith, and Choi}]{lu-etal-2022-neurologic}
Ximing Lu, Sean Welleck, Peter West, Liwei Jiang, Jungo Kasai, Daniel Khashabi, Ronan Le~Bras, Lianhui Qin, Youngjae Yu, Rowan Zellers, Noah~A. Smith, and Yejin Choi. 2022.
\newblock \href {https://doi.org/10.18653/v1/2022.naacl-main.57} {{N}euro{L}ogic a*esque decoding: Constrained text generation with lookahead heuristics}.
\newblock In \emph{Proceedings of the 2022 Conference of the North American Chapter of the Association for Computational Linguistics: Human Language Technologies}, pages 780--799, Seattle, United States. Association for Computational Linguistics.

\bibitem[{Lu et~al.(2021)Lu, West, Zellers, Le~Bras, Bhagavatula, and Choi}]{lu2021neurologic}
Ximing Lu, Peter West, Rowan Zellers, Ronan Le~Bras, Chandra Bhagavatula, and Yejin Choi. 2021.
\newblock \href {https://doi.org/10.18653/v1/2021.naacl-main.339} {{N}euro{L}ogic decoding: (un)supervised neural text generation with predicate logic constraints}.
\newblock In \emph{Proceedings of the 2021 Conference of the North American Chapter of the Association for Computational Linguistics: Human Language Technologies}, pages 4288--4299, Online. Association for Computational Linguistics.

\bibitem[{Mao et~al.(2021)Mao, Ma, Lei, Han, and Ren}]{mao2021extract}
Yuning Mao, Wenchang Ma, Deren Lei, Jiawei Han, and Xiang Ren. 2021.
\newblock \href {https://doi.org/10.18653/v1/2021.emnlp-main.413} {Extract, denoise and enforce: Evaluating and improving concept preservation for text-to-text generation}.
\newblock In \emph{Proceedings of the 2021 Conference on Empirical Methods in Natural Language Processing}, pages 5063--5074, Online and Punta Cana, Dominican Republic. Association for Computational Linguistics.

\bibitem[{Miao et~al.(2019)Miao, Zhou, Mou, Yan, and Li}]{miao2019cgmh}
Ning Miao, Hao Zhou, Lili Mou, Rui Yan, and Lei Li. 2019.
\newblock \href {https://doi.org/10.1609/aaai.v33i01.33016834} {{CGMH:} constrained sentence generation by metropolis-hastings sampling}.
\newblock In \emph{The Thirty-Third {AAAI} Conference on Artificial Intelligence, {AAAI} 2019, The Thirty-First Innovative Applications of Artificial Intelligence Conference, {IAAI} 2019, The Ninth {AAAI} Symposium on Educational Advances in Artificial Intelligence, {EAAI} 2019, Honolulu, Hawaii, USA, January 27 - February 1, 2019}, pages 6834--6842. {AAAI} Press.

\bibitem[{Michel et~al.(2019)Michel, Levy, and Neubig}]{NEURIPS2019_2c601ad9}
Paul Michel, Omer Levy, and Graham Neubig. 2019.
\newblock \href {https://proceedings.neurips.cc/paper_files/paper/2019/file/2c601ad9d2ff9bc8b282670cdd54f69f-Paper.pdf} {Are sixteen heads really better than one?}
\newblock In \emph{Advances in Neural Information Processing Systems}, volume~32. Curran Associates, Inc.

\bibitem[{Mostafazadeh et~al.(2016)Mostafazadeh, Chambers, He, Parikh, Batra, Vanderwende, Kohli, and Allen}]{mostafazadeh2016corpus}
Nasrin Mostafazadeh, Nathanael Chambers, Xiaodong He, Devi Parikh, Dhruv Batra, Lucy Vanderwende, Pushmeet Kohli, and James Allen. 2016.
\newblock A corpus and cloze evaluation for deeper understanding of commonsense stories.
\newblock In \emph{Proceedings of the 2016 Conference of the North American Chapter of the Association for Computational Linguistics: Human Language Technologies}, pages 839--849.

\bibitem[{OpenAI et~al.(2023)OpenAI, :, Achiam, Adler, Agarwal, Ahmad, Akkaya et~al.}]{openai2023gpt4}
OpenAI, :, Josh Achiam, Steven Adler, Sandhini Agarwal, Lama Ahmad, Ilge Akkaya, et~al. 2023.
\newblock \href {http://arxiv.org/abs/2303.08774} {Gpt-4 technical report}.

\bibitem[{OpenAI(2022)}]{chatgpt}
OpenAI. 2022.
\newblock \href {https://openai.com/blog/chatgpt} {Introducing chatgpt}.

\bibitem[{Papineni et~al.(2002)Papineni, Roukos, Ward, and Zhu}]{papineni2002bleu}
Kishore Papineni, Salim Roukos, Todd Ward, and Wei-Jing Zhu. 2002.
\newblock \href {https://doi.org/10.3115/1073083.1073135} {{B}leu: a method for automatic evaluation of machine translation}.
\newblock In \emph{Proceedings of the 40th Annual Meeting of the Association for Computational Linguistics}, pages 311--318, Philadelphia, Pennsylvania, USA. Association for Computational Linguistics.

\bibitem[{Post and Vilar(2018)}]{post2018fast}
Matt Post and David Vilar. 2018.
\newblock \href {https://doi.org/10.18653/v1/N18-1119} {Fast lexically constrained decoding with dynamic beam allocation for neural machine translation}.
\newblock In \emph{Proceedings of the 2018 Conference of the North {A}merican Chapter of the Association for Computational Linguistics: Human Language Technologies, Volume 1 (Long Papers)}, pages 1314--1324, New Orleans, Louisiana. Association for Computational Linguistics.

\bibitem[{Sha(2020)}]{sha2020gradient}
Lei Sha. 2020.
\newblock \href {https://doi.org/10.18653/v1/2020.emnlp-main.701} {Gradient-guided unsupervised lexically constrained text generation}.
\newblock In \emph{Proceedings of the 2020 Conference on Empirical Methods in Natural Language Processing (EMNLP)}, pages 8692--8703, Online. Association for Computational Linguistics.

\bibitem[{Silveira et~al.(2014)Silveira, Dozat, de~Marneffe, Bowman, Connor, Bauer, and Manning}]{silveira14gold}
Natalia Silveira, Timothy Dozat, Marie-Catherine de~Marneffe, Samuel Bowman, Miriam Connor, John Bauer, and Chris Manning. 2014.
\newblock \href {http://www.lrec-conf.org/proceedings/lrec2014/pdf/1089_Paper.pdf} {A gold standard dependency corpus for {E}nglish}.
\newblock In \emph{Proceedings of the Ninth International Conference on Language Resources and Evaluation ({LREC}'14)}, pages 2897--2904, Reykjavik, Iceland. European Language Resources Association (ELRA).

\bibitem[{Simonyan et~al.(2013)Simonyan, Vedaldi, and Zisserman}]{simonyan2013deep}
Karen Simonyan, Andrea Vedaldi, and Andrew Zisserman. 2013.
\newblock Deep inside convolutional networks: Visualising image classification models and saliency maps.
\newblock \emph{arXiv preprint arXiv:1312.6034}.

\bibitem[{Song et~al.(2020)Song, Wang, Yu, Zhang, Huang, Luo, Duan, and Zhang}]{song2020alignment}
Kai Song, Kun Wang, Heng Yu, Yue Zhang, Zhongqiang Huang, Weihua Luo, Xiangyu Duan, and Min Zhang. 2020.
\newblock \href {https://aaai.org/ojs/index.php/AAAI/article/view/6418} {Alignment-enhanced transformer for constraining {NMT} with pre-specified translations}.
\newblock In \emph{The Thirty-Fourth {AAAI} Conference on Artificial Intelligence, {AAAI} 2020, The Thirty-Second Innovative Applications of Artificial Intelligence Conference, {IAAI} 2020, The Tenth {AAAI} Symposium on Educational Advances in Artificial Intelligence, {EAAI} 2020, New York, NY, USA, February 7-12, 2020}, pages 8886--8893. {AAAI} Press.

\bibitem[{Song et~al.(2019)Song, Zhang, Yu, Luo, Wang, and Zhang}]{song2019code}
Kai Song, Yue Zhang, Heng Yu, Weihua Luo, Kun Wang, and Min Zhang. 2019.
\newblock \href {https://doi.org/10.18653/v1/N19-1044} {Code-switching for enhancing {NMT} with pre-specified translation}.
\newblock In \emph{Proceedings of the 2019 Conference of the North {A}merican Chapter of the Association for Computational Linguistics: Human Language Technologies, Volume 1 (Long and Short Papers)}, pages 449--459, Minneapolis, Minnesota. Association for Computational Linguistics.

\bibitem[{Sun et~al.(2023)Sun, Tian, Zhou, Xu, Hu, Gupta, Wieting, Peng, and Ma}]{sun2023evaluating}
Jiao Sun, Yufei Tian, Wangchunshu Zhou, Nan Xu, Qian Hu, Rahul Gupta, John~Frederick Wieting, Nanyun Peng, and Xuezhe Ma. 2023.
\newblock Evaluating large language models on controlled generation tasks.
\newblock \emph{arXiv preprint arXiv:2310.14542}.

\bibitem[{Tenney et~al.(2018)Tenney, Xia, Chen, Wang, Poliak, McCoy, Kim, Van~Durme, Bowman, Das et~al.}]{tenney2018you}
Ian Tenney, Patrick Xia, Berlin Chen, Alex Wang, Adam Poliak, R~Thomas McCoy, Najoung Kim, Benjamin Van~Durme, Samuel~R Bowman, Dipanjan Das, et~al. 2018.
\newblock What do you learn from context? probing for sentence structure in contextualized word representations.
\newblock In \emph{International Conference on Learning Representations}.

\bibitem[{Touvron et~al.(2023)Touvron, Martin, Stone, Albert, Almahairi et~al.}]{touvron2023llama}
Hugo Touvron, Louis Martin, Kevin Stone, Peter Albert, Amjad Almahairi, et~al. 2023.
\newblock \href {http://arxiv.org/abs/2307.09288} {Llama 2: Open foundation and fine-tuned chat models}.

\bibitem[{Vedantam et~al.(2015)Vedantam, Lawrence~Zitnick, and Parikh}]{vedantam2015cider}
Ramakrishna Vedantam, C~Lawrence~Zitnick, and Devi Parikh. 2015.
\newblock Cider: Consensus-based image description evaluation.
\newblock In \emph{Proceedings of the IEEE conference on computer vision and pattern recognition}, pages 4566--4575.

\bibitem[{Vuli{\'c} et~al.(2020)Vuli{\'c}, Ponti, Litschko, Glava{\v{s}}, and Korhonen}]{vulic2020probing}
Ivan Vuli{\'c}, Edoardo~Maria Ponti, Robert Litschko, Goran Glava{\v{s}}, and Anna Korhonen. 2020.
\newblock Probing pretrained language models for lexical semantics.
\newblock In \emph{Proceedings of the 2020 Conference on Empirical Methods in Natural Language Processing (EMNLP)}, pages 7222--7240.

\bibitem[{Wang et~al.(2023{\natexlab{a}})Wang, Li, Dai, Chen, Zhou, Meng, Zhou, and Sun}]{wang2023label}
Lean Wang, Lei Li, Damai Dai, Deli Chen, Hao Zhou, Fandong Meng, Jie Zhou, and Xu~Sun. 2023{\natexlab{a}}.
\newblock Label words are anchors: An information flow perspective for understanding in-context learning.
\newblock \emph{arXiv preprint arXiv:2305.14160}.

\bibitem[{Wang et~al.(2023{\natexlab{b}})Wang, Yu, Zeng, Yang, Wang, Chen, Jiang, Xie, Wang, Xie et~al.}]{wang2023pandalm}
Yidong Wang, Zhuohao Yu, Zhengran Zeng, Linyi Yang, Cunxiang Wang, Hao Chen, Chaoya Jiang, Rui Xie, Jindong Wang, Xing Xie, et~al. 2023{\natexlab{b}}.
\newblock Pandalm: An automatic evaluation benchmark for llm instruction tuning optimization.
\newblock \emph{arXiv preprint arXiv:2306.05087}.

\bibitem[{Wang et~al.(2022)Wang, Mishra, Alipoormolabashi, Kordi, Mirzaei, Naik, Ashok, Dhanasekaran, Arunkumar, Stap et~al.}]{wang2022super}
Yizhong Wang, Swaroop Mishra, Pegah Alipoormolabashi, Yeganeh Kordi, Amirreza Mirzaei, Atharva Naik, Arjun Ashok, Arut~Selvan Dhanasekaran, Anjana Arunkumar, David Stap, et~al. 2022.
\newblock Super-naturalinstructions: Generalization via declarative instructions on 1600+ nlp tasks.
\newblock In \emph{Proceedings of the 2022 Conference on Empirical Methods in Natural Language Processing}, pages 5085--5109.

\bibitem[{Wang et~al.(2021{\natexlab{a}})Wang, Wood, Wan, Dras, and Johnson}]{wang2021mention}
Yufei Wang, Ian Wood, Stephen Wan, Mark Dras, and Mark Johnson. 2021{\natexlab{a}}.
\newblock \href {https://doi.org/10.18653/v1/2021.acl-long.9} {Mention flags ({MF}): Constraining transformer-based text generators}.
\newblock In \emph{Proceedings of the 59th Annual Meeting of the Association for Computational Linguistics and the 11th International Joint Conference on Natural Language Processing (Volume 1: Long Papers)}, pages 103--113, Online. Association for Computational Linguistics.

\bibitem[{Wang et~al.(2021{\natexlab{b}})Wang, Xu, Hu, Tao, Wan, Dras, Johnson, and Jiang}]{wang2021neural}
Yufei Wang, Can Xu, Huang Hu, Chongyang Tao, Stephen Wan, Mark Dras, Mark Johnson, and Daxin Jiang. 2021{\natexlab{b}}.
\newblock Neural rule-execution tracking machine for transformer-based text generation.
\newblock \emph{Advances in Neural Information Processing Systems}, 34.

\bibitem[{Yao et~al.(2023)Yao, Chen, Hanjie, Yang, and Narasimhan}]{yao2023collie}
Shunyu Yao, Howard Chen, Austin~W Hanjie, Runzhe Yang, and Karthik Narasimhan. 2023.
\newblock Collie: Systematic construction of constrained text generation tasks.
\newblock \emph{arXiv preprint arXiv:2307.08689}.

\bibitem[{Zhao et~al.(2023)Zhao, Zhou, Li, Tang, Wang, Hou, Min, Zhang, Zhang, Dong et~al.}]{zhao2023survey}
Wayne~Xin Zhao, Kun Zhou, Junyi Li, Tianyi Tang, Xiaolei Wang, Yupeng Hou, Yingqian Min, Beichen Zhang, Junjie Zhang, Zican Dong, et~al. 2023.
\newblock A survey of large language models.
\newblock \emph{arXiv preprint arXiv:2303.18223}.

\bibitem[{Zheng et~al.(2023)Zheng, Chiang, Sheng, Zhuang, Wu, Zhuang, Lin, Li, Li, Xing et~al.}]{zheng2023judging}
Lianmin Zheng, Wei-Lin Chiang, Ying Sheng, Siyuan Zhuang, Zhanghao Wu, Yonghao Zhuang, Zi~Lin, Zhuohan Li, Dacheng Li, Eric Xing, et~al. 2023.
\newblock Judging llm-as-a-judge with mt-bench and chatbot arena.
\newblock \emph{arXiv preprint arXiv:2306.05685}.

\bibitem[{Zhou et~al.(2023{\natexlab{a}})Zhou, Lu, Mishra, Brahma, Basu, Luan, Zhou, and Hou}]{zhou2023instruction}
Jeffrey Zhou, Tianjian Lu, Swaroop Mishra, Siddhartha Brahma, Sujoy Basu, Yi~Luan, Denny Zhou, and Le~Hou. 2023{\natexlab{a}}.
\newblock Instruction-following evaluation for large language models.
\newblock \emph{arXiv preprint arXiv:2311.07911}.

\bibitem[{Zhou et~al.(2023{\natexlab{b}})Zhou, Jiang, Wilcox, Cotterell, and Sachan}]{zhou2023controlled}
Wangchunshu Zhou, Yuchen~Eleanor Jiang, Ethan Wilcox, Ryan Cotterell, and Mrinmaya Sachan. 2023{\natexlab{b}}.
\newblock Controlled text generation with natural language instructions.
\newblock \emph{arXiv preprint arXiv:2304.14293}.

\end{thebibliography}
\bibliographystyle{acl_natbib}

\appendix
\clearpage

\begin{table*}[t!]
    \centering
    \small
    \begin{tabular}{lp{0.8\textwidth}}
        \toprule
        Category & Keywords \\
        \midrule
        Nouns & cat, tree, book, phone, car, dog, chair, flower, house, computer, sun, water, bird, music, shoe, sky, city, mountain, river, beach, table, food, friend, love, work \\
        Verbs & run, eat, sleep, talk, walk, jump, sing, dance, write, read, play, study, think, work, swim, drive, fly, laugh, cry, climb, cook, drink, smile, fight, help \\
        Adjectives & happy, sad, beautiful, ugly, kind, cruel, smart, dumb, funny, serious, young, old, rich, poor, fast, slow, tall, short, fat, thin, strong, weak, bright, dark, clean \\
        Adverbs & quickly, slowly, happily, sadly, loudly, quietly, well, badly, carefully, carelessly, easily, hard, softly, roughly, gently, firmly, loosely, together, apart, always, never, sometimes, rarely, usually, often \\
        \bottomrule
    \end{tabular}
    \caption{The keyword set used in this paper for structural constraint.}
    \label{tab: random keywords}
\end{table*}

\begin{table*}[t!]
    \centering
    \small
    \begin{tabular}{lp{0.8\textwidth}}
        \toprule
        Category & Prompt Template \\
        \midrule
        \texttt{Keyword} & Generate a sentence with keywords: \{Keywords\} \\
        & Write a sentence with keywords: \{Keywords\} \\
        & Generate a sentence that must include the following keywords: \{Keywords\} \\
        & Please write a sentence with these keywords: \{Keywords\} \\
        & Use these keywords to generate a sentence: \{Keywords\} \\
        \midrule
        \texttt{InSen}  & Generate a story where the \{ordinal\} sentence of the story must contain the word "\{word\}" \\
        & Write a story where the \{ordinal\} sentence of the story must contain the word "\{word\}" \\
        & Generate a paragraph where the \{ordinal\} sentence of the paragraph must include the word "\{word\}" \\
        & Generate a paragraph, the \{ordinal\} sentence of the paragraph must contain the word "\{word\}" \\
        & Please write a paragraph, the \{ordinal\} sentence of the paragraph must include the word "\{word\}" \\
        \midrule
        \texttt{Order}  & Generate a sentence which contains "\{word1\}" and "\{word2\}", the word "\{word1\}" must come before "\{word2\}" in the sentence' \\
        & Write a sentence which contains "\{word1\}" and "\{word2\}", the word "\{word1\}" must come before "\{word2\}" in this sentence \\
        & Generate a sentence which includes "\{word1\}" and "\{word2\}", and "\{word1\}" must come before "\{word2\}" in the sentence \\
        & Generate a sentence which contains "\{word1\}" and "\{word2\}", the word "\{word1\}" is before "\{word2\}" in this sentence \\
        & Write a sentence which contains "\{word1\}" and "\{word2\}", the word "\{word1\}" is before "\{word2\}" in the sentence \\
        \midrule
        \texttt{WordCount} & Generate a sentence with exactly \{length\} words \\
        & Generate a sentence using exactly \{length\} words \\
        & Write a sentence with exactly \{length\} words \\
        & Generate a sentence using exactly \{length\} words \\
        & Write a sentence with exactly \{length\} words \\
        \midrule
        \texttt{SentCount} & Generate a paragraph with exactly \{length\} sentences \\
            & Generate a story using exactly \{length\} sentences \\
            & Write a paragraph with exactly \{length\} sentences \\
            & Generate a story using exactly \{length\} sentences \\
            & Write a story with exactly \{length\} sentences \\
        \midrule
        \texttt{Rel} & Question: Generate a sentence with keywords: "\{Example Keyword1\}" and "\{Example Keyword2\}", where the dependency relation between "\{Example Keyword1\}" and "\{Example Keyword2\}" is "\{Relation\}". \\
        & Answer: \{Example Sentence\} \\
        & Question: Generate a sentence with keywords: "\{Keyword1\}" and "\{Keyword2\}", where the dependency relation between "\{Keyword1\}" and "\{Keyword2\}" is "\{Relation\}". \\
        & Answer: \\
        \bottomrule
    \end{tabular}
    \caption{The prompt template used in this paper for evaluation.}
    \label{tab: template}
\end{table*}

\section{Dataset Introduction}
\label{appendix: dataset}

\subsection{Lexical Constraint}
Following the settings of previous works~\cite{lu2021neurologic,lu-etal-2022-neurologic}, we adopt the \textsc{CommonGen}\footnote{\url{http://inklab.usc.edu/CommonGen/}}~\cite{lin2020commongen} dataset for our evaluation. The \textsc{CommonGen} dataset is a widely-used constrained text generation challenge that can explicitly evaluate the LLMs for the ability of generative commonsense reasoning. The input keyword set $X$ of the \textsc{CommonGen} dataset is a set of common concepts (e.g. ``guitar'', ``sit", ``front", ``microphone"). The output is a coherent sentence containing these concepts (e.g. ``A man sits in front of a shop sings so well through a microphone and plays amazing music with his guitar.''). The \textsc{CommonGen} dataset includes 35,141 concept sets with 32,651 training samples, 993 validation samples and 1,497 test samples.

In this work, we evalute the LLMs under a \textit{zero-shot} setting. We only use the test set of the \textsc{CommonGen} dataset for evaluation, which has an average size of the concept sets of 4.04, without fine-tuning LLMs on the training set. This zero-shot evaluation can directly test the model's ability to satisfy the lexical constraints. For each sample of lexical constraints, we randomly assign a prompt template presented in Table~\ref{tab: template}.

\subsection{Structural Constraint}
An uncomplicated method to create an evaluation dataset involves utilizing story generation benchmarks such as ROCStories~\cite{mostafazadeh2016corpus}. Nevertheless, these publicly available benchmarks might have already been incorporated into the training data of language models during the pretraining stage, making it unsuitable to employ these datasets. Therefore, we manually constructed the test samples. Firstly, we ask GPT-3.5 to provide a keyword set $\mathcal{D}_w$ including 20 verbs, 20 nouns, 20 adjectives and 20 adverbs. As shown in Table~\ref{tab: random keywords}, these keywords are all commonly-used words. For constraint \texttt{InSen}$(w, y^k)$, we randomly choose $k\in [1,10]\ (k \in \mathbb{N}^+)$ and $w \in \mathcal{D}_w$. For constraint \texttt{Order}$(w_i, w_j)$, we randomly choose $w_i, w_j \in \mathcal{D}_w$ and make sure $w_i \neq w_j$. For constraint \texttt{WordCount}$(l)$, we randomly choose $l\in[5, 30]$. For constraint \texttt{SentCount}$(l)$, we randomly choose $l\in[5, 20]$. We construct 1000 samples for each category of structural constraints. For each sample of all structural constraints, we randomly assign a prompt template presented in Table~\ref{tab: template}.

\subsection{Relation Constraint}
We then construct the dataset for the evaluation of relation constraint from the English-EWT~\cite{silveira14gold} corpus\footnote{\url{https://universaldependencies.org/}}, which contains 16,621 sentences with human annotations of dependency relations. In contrast to~\citet{NEURIPS2022_ab63a1a3}, we provide a single relation triplet during testing instead of multiple ones. Furthermore, we categorize these triplets based on distinct types of dependency relations and sample 100 instances for each category to form the English-EWT corpus. Categories with fewer than 100 instances are filtered out, resulting in a final compilation of 25 categories comprising a cumulative total of 2500 test instances.

We use an in-context learning prompt to evaluate relation constraints, because dependency relation needs to be first defined by an example sentence. The template is presented in Table~\ref{tab: template}.

\section{Evaluation Metric}
\label{appendix: metric}

In the selection of evaluation metrics for constrained text generation tasks, we primarily consider two categories of metrics: one for assessing the degree of constraint fulfillment (e.g., accuracy, correlation) and another for evaluating the quality of the generated text itself (e.g., perplexity). 

\subsection{Lexical Constraint}
To assess the efficacy of constrained text generation with lexical constraints, previous studies~\cite{lu2021neurologic,lu-etal-2022-neurologic} commonly employ various automatic evaluation metrics, including BLEU~\cite{papineni2002bleu}, ROUGE~\cite{lin2004rouge}, METEOR~\cite{banerjee2005meteor}, CIDEr~\cite{vedantam2015cider}, and SPICE~\cite{anderson2016spice}. These metrics quantify the similarity between the generated text and reference samples. Nevertheless, these automatic metrics are not a reasonable way to evaluate the performance because constrained text generation is an open-ended task, lacking definitive reference outputs. Thus, in this paper, we solely present the following metrics: Accuracy (the percentage of sentences fulfilling all lexical constraints), Coverage (the percentage of input keywords that are present in lemmatizatized outputs), BLEU-4, ROUGE-L, and GPT-2 PPL.

\begin{table*}[t!]
    \centering
    \small
    
    \begin{tabular}{l|p{0.75\textwidth}}
    \toprule
    \textbf{Input} & Generate a sentence with keywords: ``\textcolor{blue}{improvise}", ``\textcolor{blue}{barrel}", ``\textcolor{blue}{transport}", ``\textcolor{red}{work}", ``\textcolor{blue}{tool}". \\
    GPT-3.5 & The \textcolor{red}{workers} had to \textcolor{blue}{improvise} a \textcolor{blue}{tool} to \textcolor{blue}{transport} the heavy \textcolor{blue}{barrel} to the other side of the factory. \\
    \midrule
    \textbf{Input} & Generate a sentence with keywords: ``\textcolor{blue}{pile}", ``\textcolor{red}{fall}", ``\textcolor{blue}{ground}", ``\textcolor{blue}{jump}", ``\textcolor{blue}{snow}". \\
    GPT-4 & After a heavy \textcolor{red}{snowfall}, the children couldn't resist the urge to \textcolor{blue}{jump} into the massive \textcolor{blue}{pile} of \textcolor{blue}{snow} that had accumulated on the \textcolor{blue}{ground}. \\
    \bottomrule
    \end{tabular}
    \caption{Some corner cases for GPT-3.5 and GPT-4.}
    \label{tab: corner case}
\end{table*}

\subsection{Structural Constraint}

Following~\citet{wang2021neural}, for all of structural constraints, we report the accuracy, the ratio of model outputs that completely satisfy the given constraints. Apart from that, we use some different automatic evaluation metrics according to different constraint types. 

For constraint \texttt{InSen}$(w, y^k)$, we also report the accuracy under error bars $\pm 1$, $\pm 2$ and $\pm 3$. Formally, for $m$ constraints $\{$\texttt{InSen}$(w_i, y_i^{k_i})\}_{i=1}^m$:
\begin{equation}
\small
\text{ACC}(\pm n)=\frac1m\sum_{i=1}^m \mathds{1}[w_i\in \text{sentence } k_i-n \text{ to } k_i+n \text{ of } y_i].
\end{equation}

For \texttt{WordCount} and \texttt{SentCount}, we report Pearson correlation, Kendall-Tau correlation, accuracy and GPT-2 PPL.


\subsection{Relation Constraint}
In this paper, we use a widely-used dependency parser provided by \texttt{spaCy}\footnote{\url{https://explosion.ai/blog/ud-benchmarks-v3-2\#project}} to determine the relation between $h$ and $t$ in the output. Similar to~\citet{NEURIPS2022_ab63a1a3}, we adopt the unlabeled/labeled/word coverage (UC/LC/WC) as the metrics for evaluation. LC (Accuracy) denotes the proportion of instances where the relation between $h$ and $t$ is accurately identified. UC refers to the proportion of instances where there exists a dependency relation between $h$ and $t$, but the relation type is incorrect. WC is analogous to the lexical constraint coverage. 


\section{Experimental Details}
\label{appendix: experimental}

\subsection{Generation Configuration}
For constrained text generation evaluation, we use a temperature of 0.8 and a top-p of 0.95 on open-source LLMs with 5 random seeds. For GPTs, we use a temperature of 0 to obtain deterministic results. In terms of constraint consistency evaluation in Section~\ref{ssec: cons}, we also use a temperature of 0. The open-source LLMs inference is speeded up by the vLLM package on a single A40 GPU with 48G memory.

\subsection{Probing Configuration}
For the probing experiment in Section~\ref{ssec: probing}, the batch size is 128 and the learning rate is 1e-5 with Adam~\cite{kingma2014adam} optimizer. The experiment is implemented on a single A40 GPU with 48G memory.

\subsection{Attention Re-anchoring Configuration}
We choose the re-anchoring parameter $\lambda=e^{0.5}-1$. The results are not sensitive to the choice of $\lambda$.

\section{Case Study}
\label{appendix: case}

Table~\ref{tab: corner case} displays some corner cases involving GPT-3.5 and GPT-4. It can be observed that on occasion, these models may struggle when dealing with simple words. These words might appear as prefixes or suffixes in the generated output due to tokenization inherent to LLMs. However, these corner cases can be easily resolved by reiteration or by providing explicit feedback to the GPTs to rectify the outcomes.

\end{document}